\documentclass[format=acmsmall, review=false, screen=true]{acmart}
\usepackage{booktabs} 

\acmJournal{TKDD}
\acmVolume{0}
\acmNumber{1}
\acmArticle{00}
\acmYear{2018}
\acmMonth{8}

\setcopyright{acmcopyright}

\acmDOI{0000001.0000001}

\usepackage{array}
\newcolumntype{P}[1]{>{\centering\arraybackslash}p{#1}}
\newcolumntype{M}[1]{>{\centering\arraybackslash}m{#1}}
\newcolumntype{R}[1]{>{\arraybackslash}m{#1}}

\definecolor{orange}{rgb}{1,0.5,0}
\definecolor{graynode}{RGB}{20,20,20}
\definecolor{crimsonred}{RGB}{220,20,60}
\definecolor{darkgraynode}{gray}{0.5}
\definecolor{lightgraynode}{gray}{0.8}

\usepackage{rotate}
\usepackage{capt-of}
\usepackage{tabulary}
\usepackage{setspace}
\usepackage{amssymb}
\usepackage{pifont} 
\newcommand{\cmark}{\ding{51}}
\newcommand{\xmark}{\ding{55}}

\definecolor{gray}{RGB}{20,20,20}
\definecolor{gray}{RGB}{0.7,0.7,0.7}
\definecolor{greencm}{RGB}{0,153,0}

\newcommand{\cm}{ {\color{greencm}\normalsize\cmark}}
\newcommand{\cmgray}{ {\color{gray}\normalsize\cmark}}
\newcommand{\xm}{ {\color{red}\normalsize\xmark}}

\definecolor{plotblue}{RGB}	{30,144,255}
\definecolor{plotgreen}{RGB}	{50,205,50}
\definecolor{plotred}{RGB}	{220,20,60}

\definecolor{myyellow}{RGB}{255,255,204}
\definecolor{myred}{RGB}{255,204,204}
\definecolor{myblue}{RGB}{0,200,255}
\definecolor{mygreen}{RGB}{80,220,80}

\newcommand*\hrulefillvar[1][0.4pt]{\leavevmode\leaders\hrule height#1\hfill\kern0pt}

\usepackage{ragged2e} 

\newcommand{\etal}{\emph{et al.}}

\newcommand{\eg}{\emph{e.g.}}
\newcommand{\ie}{\emph{i.e.}}

\newtheorem{Definition}{Definition}

\usepackage{comment}
\usepackage{tabularx}

\usepackage{subfigure}
\usepackage{enumitem}

\definecolor{thedarkblue}{RGB}{0,0,120} 
\definecolor{mydarkblue}{rgb}{0,0.08,0.45} 

\usepackage{hyperref}
\hypersetup{%
    colorlinks=true,
    linkcolor=mydarkblue,
    citecolor=mydarkblue,
    filecolor=mydarkblue,
    urlcolor=mydarkblue}
    
\usepackage{microtype}
\usepackage{graphicx}

\usepackage{balance}
\usepackage{tabularx}

\usepackage{booktabs}
\usepackage{tabularx}
\usepackage{comment}

\providecommand{\mat}[1]{\boldsymbol{\mathrm{#1}}}%
\renewcommand{\vec}[1]{\boldsymbol{\mathrm{#1}}}

\DeclareMathOperator{\hugeE}{\mbox{\huge\raise-0.3ex\hbox{E}}}
\DeclareMathOperator{\p}{\mathbb{P}}
\DeclareMathOperator{\hugep}{\mbox{\huge\raise-0.3ex\hbox{$\p$}}}

\newcommand{\RR}{\mathbb{R}}

\providecommand{\mA}{\ensuremath{\mat{A}}}

\providecommand{\mD}{\ensuremath{\mat{D}}}

\providecommand{\mH}{\ensuremath{\mat{H}}}
\providecommand{\mI}{\ensuremath{\mat{I}}}

\providecommand{\mT}{\ensuremath{\mat{T}}}
\providecommand{\mU}{\ensuremath{\mat{U}}}

\providecommand{\mW}{\ensuremath{\mat{W}}}
\providecommand{\mX}{\ensuremath{\mat{X}}}

\providecommand{\mZ}{\ensuremath{\mat{Z}}}

\providecommand{\va}{\ensuremath{\vec{a}}}
\providecommand{\vb}{\ensuremath{\vec{b}}}

\providecommand{\ve}{\ensuremath{\vec{e}}}

\providecommand{\vg}{\ensuremath{\vec{g}}}
\providecommand{\vh}{\ensuremath{\vec{h}}}

\providecommand{\vr}{\ensuremath{\vec{r}}}
\providecommand{\vs}{\ensuremath{\vec{s}}}
\providecommand{\vt}{\ensuremath{\vec{t}}}

\providecommand{\vx}{\ensuremath{\vec{x}}}

\providecommand{\vz}{\ensuremath{\vec{z}}}

\providecommand{\N}{\ensuremath{\Gamma}} 

\begin{document}
\title{Attention Models in Graphs: A Survey}


\author{John Boaz Lee}
\affiliation{%
  \institution{WPI}
  \state{MA}
  \country{USA}
}
\email{jtlee@wpi.edu}

\author{Ryan A. Rossi}
\orcid{1234-5678-9012-3456}
\affiliation{%
  \institution{Adobe Research}
  \city{San Jose}
  \state{CA}
  \country{USA}
}
\email{rrossi@adobe.com}

\author{Sungchul Kim}
\affiliation{%
  \institution{Adobe Research}
  \city{San Jose}
  \state{CA}
  \country{USA}
}
\email{sukim@adobe.com}

\author{Nesreen K. Ahmed}
\affiliation{%
  \institution{Intel Labs}
  \city{Santa Clara}
  \state{CA}
  \country{USA}
}
\email{nesreen.k.ahmed@intel.com}

\author{Eunyee Koh}
\affiliation{%
  \institution{Adobe Research}
  \city{San Jose}
  \state{CA}
  \country{USA}
}
\email{eunyee@adobe.com}

%

\renewcommand\shortauthors{Lee, J. et al}

\begin{abstract}
Graph-structured data arise naturally in many different application domains. By representing data as graphs, we can capture entities (\textit{i.e.}, nodes) as well as their relationships (\textit{i.e.}, edges) with each other. Many useful insights can be derived from graph-structured data as demonstrated by an ever-growing body of work focused on graph mining. However, in the real-world, graphs can be both large -- with many complex patterns -- and noisy which can pose a problem for effective graph mining. An effective way to deal with this issue is to incorporate ``attention'' into graph mining solutions. An attention mechanism allows a method to focus on task-relevant parts of the graph, helping it to make better decisions. In this work, we conduct a comprehensive and focused survey of the literature on the emerging field of graph attention models. We introduce three intuitive taxonomies to group existing work. These are based on problem setting (type of input and output), the type of attention mechanism used, and the task (\textit{e.g.}, graph classification, link prediction, \textit{etc.}). We motivate our taxonomies through detailed examples and use each to survey competing approaches from a unique standpoint. Finally, we highlight several challenges in the area and discuss promising directions for future work.
\end{abstract}




\maketitle

\section{Introduction}

Data which can be naturally modeled as graphs are found in a wide variety of domains including the world-wide web~\cite{Albert99}, bioinformatics~\cite{Pei05}, neuroscience~\cite{Lee17}, chemoinformatics~\cite{Duvenaud15}, social networks~\cite{Backstrom11}, scientific citation and collaboration~\cite{Liu17}, urban computing~\cite{Zheng14}, recommender systems~\cite{Deng17}, sensor networks~\cite{Aggarwal17}, epidemiology~\cite{Moslonka11}, anomaly and fraud analysis~\cite{Akoglu15}, and ecology~\cite{Allesina05}. For instance, interactions between users on a social network can be captured using a graph where the nodes represent users and links denote user interaction and/or friendship~\cite{Backstrom11}. On the other hand, in chemoinformatics, we can build molecular graphs by treating the atoms as nodes and the bonds between atoms as edges~\cite{Duvenaud15}.

A large body of work -- which we broadly categorize as the field of graph mining~\cite{Aggarwal10} -- has emerged that focuses on gaining insights from graph data. Many interesting and important problems have been studied in this area. These include graph classification~\cite{Duvenaud15}, link prediction~\cite{Sun11},  community detection~\cite{Newman02}, functional brain network discovery~\cite{Bai17}, node classification and clustering~\cite{Perrozi15}, and influence maximization~\cite{He14} with new problems constantly being proposed.

In the real-world, however, graphs can be both structurally large and complex~\cite{AhmedTKDD,Leskovec06} as well as noisy~\cite{He14}. These pose a significant challenge to graph mining techniques, particularly in terms of performance~\cite{ahmed17streams}.

\begin{figure}[t]
\centering
\includegraphics[width=0.55\linewidth]{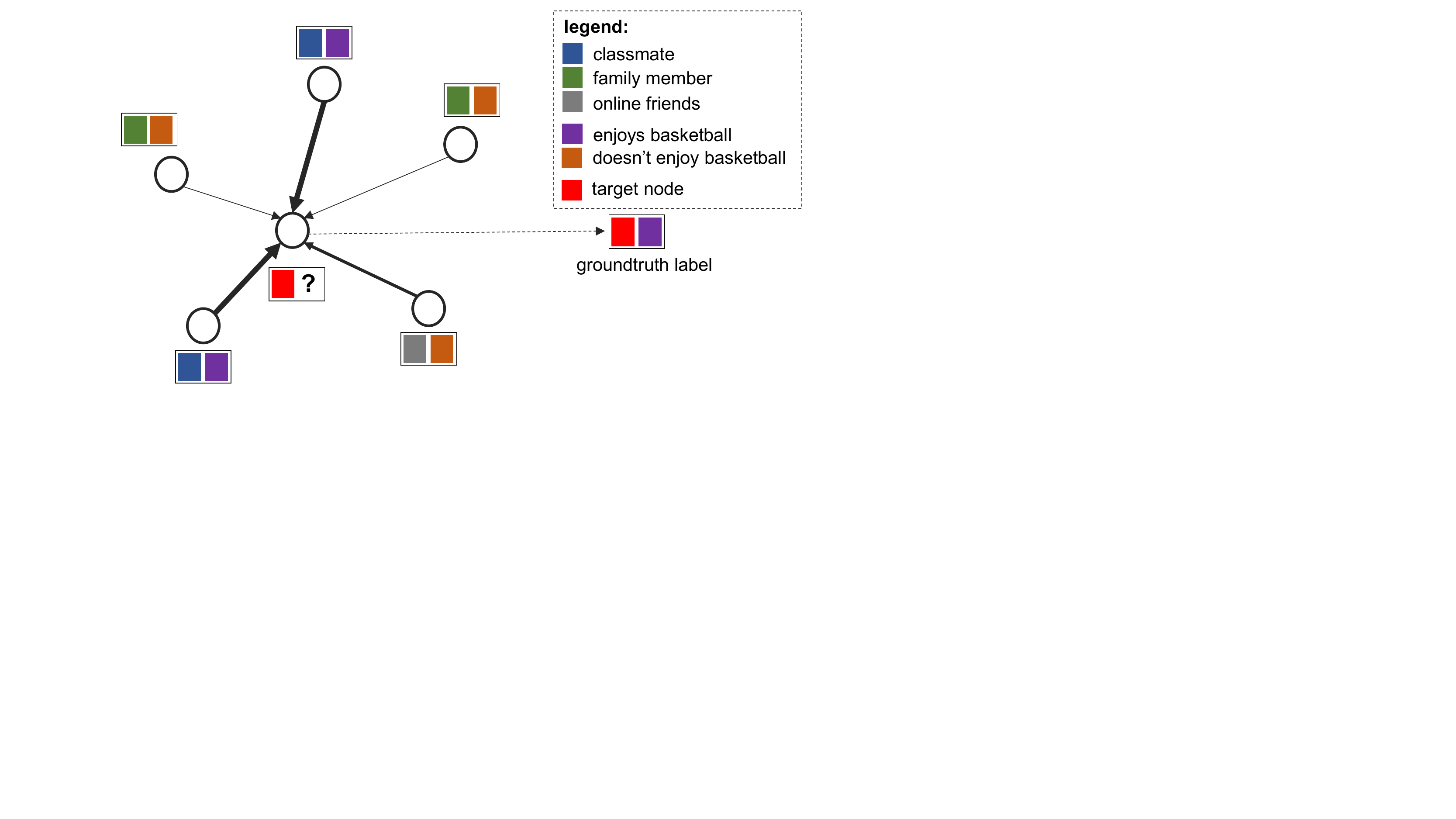}
\caption{Attention is used to assign importance to each type of neighbor. The link size denotes how much attention we want to apply to each neighbor. In this example, we see that by using attention to focus on a node's classmates, we can better predict the kind of activity the target is interested in. Best viewed with color.}
\label{fig:att-motiv}
\end{figure}

Various techniques have been proposed to address this issue. For instance, the method proposed by~\cite{Wu16} utilizes multiple views of the same graph to improve classification performance while~\cite{Zhang16} leverages auxiliary non-graph data as side views under similar motivation. Another popular technique involves the identification of task-relevant or discriminative subgraphs~\cite{Shi12,Zhu12}.

Recently, a new approach has emerged to address the above-mentioned problem and this is by incorporating \textit{attention} into graph mining solutions. An attention mechanism aids a model by allowing it to "focus on the most relevant parts of the input to make decisions"~\cite{Veli18}. Attention was first introduced in the deep learning community to help models attend to important parts of the data~\cite{Mnih14, Bah15}. The mechanism has been successfully adopted by models solving a variety of tasks. Attention was used by~\cite{Mnih14} to take glimpses of relevant parts of an input image for image classification; on the other hand,~\cite{Xu15} used attention to focus on important parts of an image for the image captioning task. Meanwhile~\cite{Bah15} utilized attention for the machine translation task by assigning weights which reflected the importance of different words in the input sentence when generating corresponding words in the output sentence. Finally, attention has also been used for both the image~\cite{Yu16} as well as the natural language~\cite{Kumar16} question answering tasks. However, most of the work involving attention has been done in the computer vision or natural language processing domains.

More recently, there has been a growing interest in attention models for graphs and various techniques have been proposed~\cite{Feng16, Choi17, Ma17, Han18, Lee18, Veli18}. Although attention is defined in slightly different ways in all these papers, the competing approaches do share common ground in that attention is used to allow the model to focus on task-relevant parts of the graph. We discuss and define, more precisely, the main strategies used by these methods to apply attention to graphs in Section~\ref{sec:mechanism}. Figure~\ref{fig:att-motiv} shows a motivating example of when attention can be useful in a graph setting. In particular, the main advantages of using attention on graphs can be summarized as follows:
\begin{enumerate}
\item Attention allows the model to avoid or ignore noisy parts of the graph, thus improving the signal-to-noise (SNR) ratio~\cite{Mnih14,Lee18}.
\item Attention allows the model to assign a relevance score to elements in the graph (for instance, the different neighbors in a node's neighborhood) to highlight elements with the most task-relevant information, further improving SNR~\cite{Veli18}.
\item Attention also provides a way for us to make a model's results more interpretable~\cite{Veli18, Choi17}. For example, by analyzing a model's attention over different components in a medical ontology graph we can identify the main factors that contribute to a particular medical condition~\cite{Choi17}.
\end{enumerate}

In this paper, we conduct a comprehensive and focused review of the literature on graph attention models. To the best of our knowledge, this is the first work on this topic. We introduce three different taxonomies to group existing work into intuitive categories. We then motivate our taxonomies through detailed examples and use each to survey competing approaches from a particular standpoint. In particular, we group the existing work by problem setting (defined by the type of input graph and the primary problem output), by the kind of attention that is used, and by the task (\textit{e.g.}, node classification, graph classification, \textit{etc.}). 

\begin{figure}[t]
\centering
\includegraphics[width=0.6\linewidth]{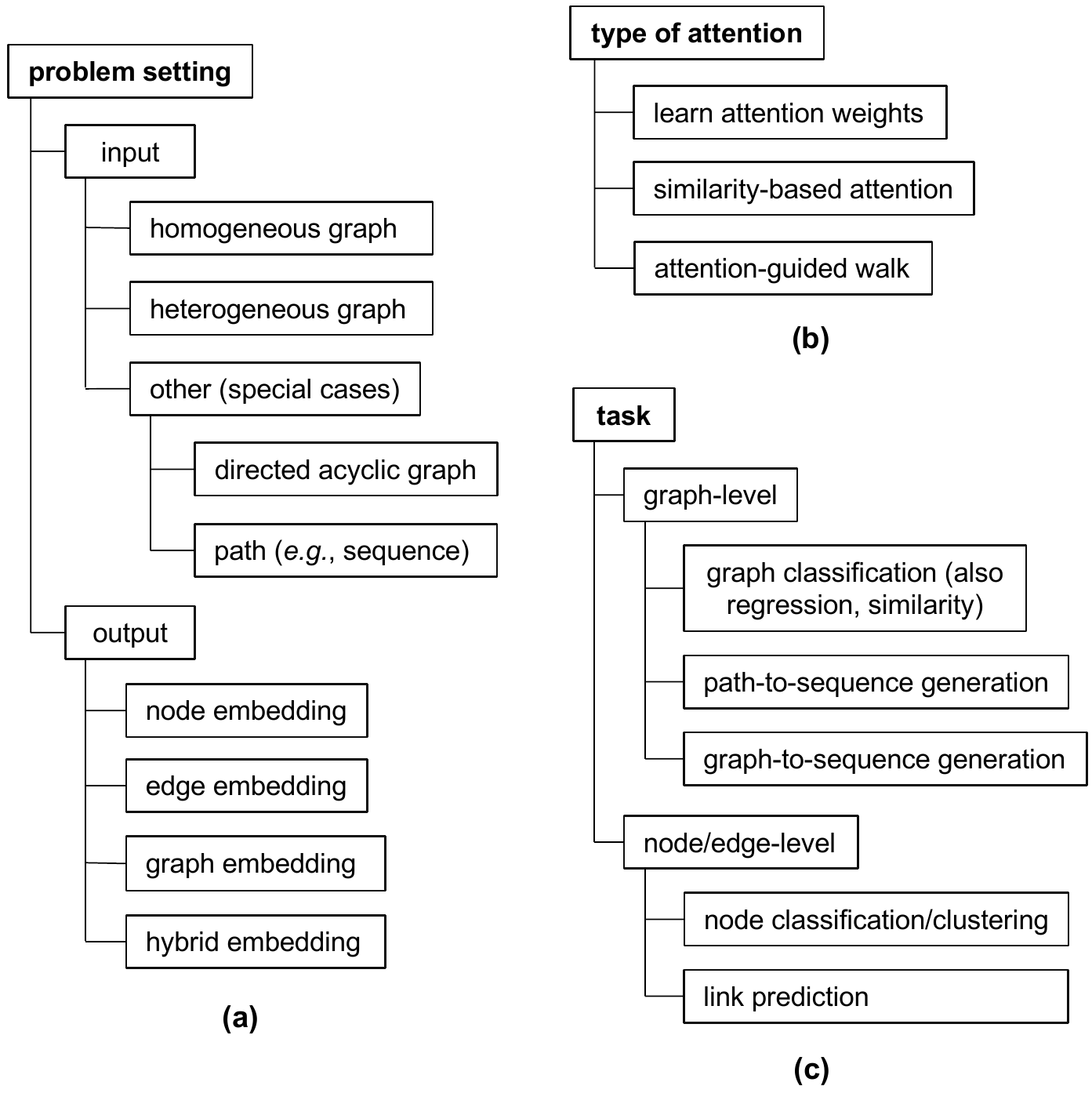}
\caption{Proposed taxonomies to group graph attention models based on (a) problem setting, (b) type of attention used, and (c) task or problem.}
\label{fig:tax}
\vspace{-5mm}
\end{figure}

In previous work, different kinds of graphs (\textit{e.g.}, homogeneous graphs, heterogeneous graphs, directed acyclic graphs, \textit{etc.}) have been studied with different properties (\textit{e.g.}, attributed, weighted or unweighted, directed or undirected) and different outputs (\textit{e.g.}, node embedding, link embedding, graph embedding). The first taxonomy allows us to survey the field from this perspective. The second taxonomy tackles the main strategies that have been proposed for applying attention in graphs. We then introduce a final taxonomy that groups the methods by application area; this shows the reader what problems have already been tackled while, perhaps more importantly, revealing important graph-based problems where attention models have yet to be applied. Figure~\ref{fig:tax} shows the proposed taxonomies.

Additionally, we also summarize the challenges that have yet to be addressed in the area of graph attention and provide promising directions for future work. 

\subsection{Main contributions}
The main contributions of this work are as follows:
\begin{enumerate}
\item We introduce three intuitive taxonomies for categorizing various graph attention models and survey existing methods using these taxonomies. To the best of our knowledge, this is the first survey on the important field of graph attention.
\item We motivate each taxonomy by discussing and comparing different graph attention models from the taxonomy's perspective.
\item We highlight the challenges that have yet to be addressed in the area of graph attention.
\item We also provide suggestions on potential areas for future work in this emerging field. 
\end{enumerate}

\subsection{Scope of this article}
In this article, we focus on examining and categorizing various techniques that apply attention to graphs (we give a general definition of attention in Sec.~\ref{sec:formulation}). Most of these methods take graphs as input and solve some graph-based problem such as link prediction~\cite{Sun11}, graph classification/regression~\cite{Duvenaud15}, or node classification~\cite{gcn}. However, we also consider methods that apply attention to graphs although the graph is only one of several types of input to the problem.  

We do \textit{not} attempt to survey the vast field of general graph-based methods that do not explicitly apply attention, multiple work have already been done on this with each having a particular focus~\cite{emb-survey-tkde, survey-freqsubgraph, survey-link}.

\subsection{Organization of the survey}
The rest of this survey is organized as follows. We start by introducing useful notations and definitions in Section~\ref{sec:formulation}. We then use Sections~\ref{sec:attention-based-node-embedding} through~\ref{sec:attention-based-hybrid-embedding} to discuss related work using our main taxonomy (Fig.~\ref{fig:tax}a). We organized Sections~\ref{sec:attention-based-node-embedding}-\ref{sec:attention-based-hybrid-embedding} such that the methods in existing work are grouped by the main type of embedding they calculate (\textit{e.g.}, node embedding, edge embedding, graph embedding, or hybrid embedding); these methods are then further divided by the type of graphs they support. In Sections~\ref{sec:mechanism} and~\ref{sec:tasks}, we switch to a different perspective and use the remaining taxonomies (Fig.~\ref{fig:tax}b and Fig.~\ref{fig:tax}c) to guide the discussion. We then discuss challenges as well as interesting opportunities for future work in Section~\ref{sec:discussion-challenges}. Finally, we conclude the survey in Section~\ref{sec:conclusion}.

\section{Problem Formulation} \label{sec:formulation}
In this section, we define the different types of graphs that appear in the discussion; we also introduce related notations.

\begin{Definition}[Homogeneous graph] \label{def:homo-graph}
Let $G=(V,E)$ be a graph where $V$ is the set of nodes (vertices) and $E$ is the set of edges between the nodes in $V$. 
\end{Definition}

Further, let $\mA = \big[ A_{ij}\big]$ be the $n \times n$, for $n = |V|$, adjacency matrix of $G$ where $A_{ij}=1$ if there exists $(v_i,v_j) \in E$ and $A_{ij}=0$ otherwise. When $\mA$ is a binary matrix (\ie, $A_{ij} \in \{0,1\}$), we consider the graph to be \textit{unweighted}. Note that $\mA$ may also encode edge weights; given a weight $w \in \mathbb{R}$ for $(v_i,v_j) \in E$, $A_{ij}=w$. In this case, the graph is said to be \textit{weighted}. Also, if $A_{ij} = A_{ji}$, for any $1 \leq i, j \leq n$, then the graph is \textit{undirected}, otherwise it is \textit{directed}.

Given an adjacency matrix $\mA$, we can construct a right stochastic matrix $\mT$ -- also called the transition matrix -- which is simply $\mA$ with rows normalized to $1$. Also, given a vertex $v \in V$, let $\N_v$ be the set of vertices in node $v$'s neighborhood (for instance, a popular definition for neighborhood is simply the one-hop neighborhood of $v$, or $\N_v = \{w | (v,w) \in E\}$). 

\begin{Definition}[Heterogeneous graph] \label{def:heter-graph}
A heterogeneous graph is defined as $G=(V,E)$ consisting of a set of node objects $V$ and a set of edges $E$ connecting the nodes in $V$. A heterogeneous graph also has a \emph{node type mapping function} $\,\theta : V \rightarrow \mathcal{T}_V$ 
and an \emph{edge type mapping function} defined as $\,\xi : E \rightarrow \mathcal{T}_E$ 
where $\mathcal{T}_V$ and $\mathcal{T}_E$ denote the set of node types and edge types, respectively.
The type of node $i$ is denoted as $\theta(i)$ (which may be an author, a paper, or a conference in a heterogeneous bibliographic network) whereas the type of edge $e = (i,j) \in E$ is denoted as $\xi(i,j)=\xi(e)$ (\eg, a co-authorship relationship, or a ``published in'' relationship).
\end{Definition}\noindent

Note a heterogeneous graph is sometimes referred to as a typed network. Furthermore, a bipartite graph is a simple heterogeneous graph with two node types and a single edge type. A homogeneous graph is simply a special case of a heterogeneous graph where $|\mathcal{T}_V| = |\mathcal{T}_E| = 1$.


\begin{Definition}[Attributed graph] \label{def:attributed-graph}
Let $G=(V,E, \mX)$ be an attributed graph where
$V$ is a set of nodes, 
$E$ is a set of edges, 
and $\mX$ is an $n \times d$ matrix of \emph{node input attributes} where each 
$\bar{\vx}_i$ (or $\mX_{i:}$) is a $d$-dimensional (row) vector of attribute values for node $v_i \in V$ and $\vx_j$ (or $\mX_{:j}$) is an $n$-dimensional vector corresponding to the $j$th attribute (column) of $\mX$.
Alternatively, $\mX$ may also be a $m \times d$ matrix of \emph{edge input attributes}.
These may be real-valued or not.
\end{Definition}\noindent

\begin{table}[t!]
\label{table:notation}
\caption{Summary of notation. Matrices are bold upright roman letters; vectors are bold lowercase letters.}
\vspace{-2mm}
\centering 
\fontsize{8}{8.5}\selectfont
\setlength{\tabcolsep}{6pt} 
\label{table:notation}
\hspace*{-2.5mm}
\def\arraystretch{1.38} 
\begin{tabularx}{.8\linewidth}{@{}rX@{}}
\toprule
$G$ & (un)directed (attributed) graph\\
$\mA$ & adjacency matrix of the graph $G=(V,E)$ \\
$\mT$ & stochastic transition matrix for $G$ constructed by normalizing over the rows in $\mA$\\
$n, m$ & number of nodes and edges in the graph \\
$k$ & number of embedding dimensions \\
$d$ & node attribute dimension \\
$\N_{i}$ & neighbors of $i$ defined using some neighborhood\\
$\N_i^{(\ell)}$ & $\ell$-neighborhood $\N_i^{(\ell)} = \{j \in V \,|\, \mathrm{dist}(i, j) \leq \ell \}$ \\
$\mathrm{dist}(i,j)$ &  shortest distance between $i$ and $j$ \\
$\mX$  & $n \times d$ input attribute matrix \\ 
$\vx$ & a $d$-dimensional feature vector\\
$x_i$ & the $i$-th element of $\vx$  \\
$\mZ$  & output node embedding matrix (or vector $\vz$ in the case of graph embeddings) \\ 
\bottomrule
\end{tabularx}
\end{table}

\begin{Definition}[Directed acyclic graph (DAG)] \label{def:dag}
A directed graph with no cycles.
\end{Definition}\noindent

The different ``classes'' of graphs defined in Definitions~\ref{def:homo-graph}-\ref{def:attributed-graph} can be directed or undirected as well as weighted or unweighted. Other classes of graphs arise from composing the distinct ``graph classes'' (Definition~\ref{def:homo-graph}-\ref{def:attributed-graph}). For instance, it is straightforward to define an attributed heterogeneous network $G=(V,E,\mathcal{T}_V, \mathcal{T}_E, \mX)$.

\begin{Definition}[Path] \label{def:path}
A path $P$ of length $L$ is defined as a sequence of \emph{unique} indices $i_{1}, i_{2}, \ldots, i_{L+1}$ such that $(v_{i_{t}}, v_{i_{t+1}}) \in E$ for all $1 \leq t \leq L$.
The length of a path is defined as the number of edges it contains.
\end{Definition}\noindent

Unlike walks that allow loops, paths do not and therefore a walk is a path iff it has no loops (all nodes are unique). Note that a path is a special kind of graph with a very rigid structure where all the nodes have at most $2$ neighbors. We now give a general, yet formal, definition of the notion of \emph{graph attention} as follows:

\begin{Definition}[Graph Attention]\label{def:graph-attention}
Given a target graph object (\eg, node, edge, graph, \textit{etc}), $v_0$ and a set of graph objects in $v_0$'s ``neighborhood'' $\{v_1, \cdots, v_{|\N_{v_0}|} \} \in \N_{v_0}$ (the neighborhood is task-specific and can mean the $\ell$-hop neighborhood of a node or something more general). Attention is defined as a function $f^\prime : \{ v_0 \} \times \N_{v_0} \rightarrow [0, 1]$ that maps each of the object in $\N_{v_0}$ to a relevance score which tells us how much attention we want to give to a particular neighbor object. Furthermore, it is usually expected that $\sum_{i=1}^{|\N_{v_0}|} f^\prime(v_0, v_i) = 1$. 
\end{Definition}\noindent

A summary of the notation used throughout the manuscript is provided in Table~\ref{table:notation}.

\newcommand\TE{\rule{0pt}{2.0ex}}
\newcommand\BE{\rule[-1.1ex]{0pt}{0pt}}

{
\newcolumntype{C}{ >{\centering\arraybackslash} m{4cm} } 
\providecommand{\rotateDeg}{90}
\setlength{\tabcolsep}{1.2pt}
\providecommand{\rotDeg}{70}
\definecolor{verylightgreennew}{RGB}	{220,255,220}
\definecolor{verylightrednew}{RGB}		{255, 230, 230}
\definecolor{verylightreddarker}{HTML} {FFCBCB} 
\definecolor{verylightrednew}{RGB}		{255, 230, 230}
\definecolor{verylightrednewlighter}{RGB}		{255, 229, 239}
\definecolor{lightgraynew}{rgb}{0.9,0.9,0.9}
\definecolor{newgray}{RGB}{0.7,0.7,0.7}
\providecommand{\cellsz}{0.34cm} 
\providecommand{\cellszlg}{0.36cm} 
\renewcommand{\cm}{{\color{greencm}\normalsize\cmark}}
\renewcommand{\cmgray}{{\color{newgray}\normalsize\cmark}}
\renewcommand{\xm}{{\color{verylightreddarker}\normalsize\xmark}}
\newcommand\BBBBB{\rule[1.6ex]{0pt}{1.6ex}}
\newcommand\BBBBBB{\rule[-1.1ex]{0pt}{0pt}} 
\newcommand{\sysName}[1]{{\sf
\BBBBBB
#1
}}
\providecommand{\cellsomewhat}{
\BBBBB
\cmgray
\cellcolor{lightgraynew}
}
\providecommand{\cellno}{
\BBBBB
\xm
\cellcolor{verylightrednew}}
\providecommand{\cellyes}{
\BBBBB
\cm
\cellcolor{verylightgreennew}
}

\newcolumntype{H}{>{\setbox0=\hbox\bgroup}c<{\egroup}@{}} 
\begin{table}[t!]
\def\arraystretch{1.2} 
\scriptsize
\caption{Qualitative and quantitative comparison of graph-based attention models.
%
}
\label{table:qual-and-quant-comparison}
\begin{minipage}{\columnwidth}
{\begin{center}
\vspace{-2mm}
\begin{tabular}
{P{2mm}
r
c 
!{\vrule width 0.8pt} 
P{\cellsz} P{\cellsz} 
P{\cellsz} P{\cellsz} 
!{\vrule width 0.6pt} 
P{\cellsz} P{\cellsz} 
P{\cellsz} P{\cellsz} 
!{\vrule width 0.6pt}
P{\cellszlg} 
P{\cellszlg} 
P{\cellszlg} 
P{\cellszlg} 
!{\vrule width 0.6pt} 
P{\cellsz} P{\cellsz} P{\cellsz} P{\cellsz} 
P{\cellsz} P{\cellsz} P{\cellsz}
!{\vrule width 0.8pt}
@{}
}


& & 
& \multicolumn{4}{c!{\vrule width 0.6pt}}{\textsc{Input}}
& \multicolumn{4}{c!{\vrule width 0.6pt}}{\textsc{Output}}
& \multicolumn{4}{c!{\vrule width 0.6pt}}{\textsc{Mechanism}}
& \multicolumn{7}{c!{\vrule width 0.8pt}}{\textsc{Task}} 
\\

& \multicolumn{1}{r}{\textbf{Method}} & 
& 
\rotatebox{\rotateDeg}{\textbf{Homogeneous graph}} & 
\rotatebox{\rotateDeg}{\textbf{Heterogeneous graph}} &
\rotatebox{\rotateDeg}{\textbf{Directed acyclic graph}} &
\rotatebox{\rotateDeg}{\textbf{Path}} &
\rotatebox{\rotateDeg}{\textbf{Node embedding}} & 
\rotatebox{\rotateDeg}{\textbf{Edge embedding}} & 
\rotatebox{\rotateDeg}{\textbf{Graph embedding}} &
\rotatebox{\rotateDeg}{\textbf{Hybrid embedding}} &
\rotatebox{\rotateDeg}{\textbf{Learn attention weights}} &
\rotatebox{\rotateDeg}{\textbf{Similarity-based attention}} &  
\rotatebox{\rotateDeg}{\textbf{Attention-guided walk}} & 
\rotatebox{\rotateDeg}{\textbf{Other}} & 
\rotatebox{\rotateDeg}{\textbf{Node/Link classification}} &
\rotatebox{\rotateDeg}{\textbf{Link prediction}} &
\rotatebox{\rotateDeg}{\textbf{Graph classification}} &
\rotatebox{\rotateDeg}{\textbf{Graph regression}} &
\rotatebox{\rotateDeg}{\textbf{Seq-to-seq generation}} &
\rotatebox{\rotateDeg}{\textbf{Graph-to-seq generation}} &
\rotatebox{\rotateDeg}{\textbf{Other}} 
\\
\noalign{\hrule height 0.8pt}

& \sysName{$\mathsf{\sf \bf Attention Walks}$}~\cite{att-node2vec}
&
& \cellyes 
& \cellno 
& \cellno
& \cellno 
& \cellyes  
& \cellno
& \cellno 
& \cellno
& \cellyes
& \cellno
& \cellyes 
& \cellno 
& \cellno
& \cellyes
& \cellno 
& \cellno
& \cellno 
& \cellno
& \cellno   
\\
\hline

& \sysName{$\mathsf{\sf \bf GAKE}$}~\cite{Feng16}
&
& \cellyes 
& \cellno 
& \cellno
& \cellno 
& \cellyes  
& \cellno
& \cellno 
& \cellno
& \cellyes
& \cellno
& \cellno 
& \cellno 
& \cellno
& \cellyes
& \cellno 
& \cellno
& \cellno 
& \cellno
& \cellno   
\\
\hline

& \sysName{$\mathsf{\sf \bf GAT}$}~\cite{Veli18}
&
& \cellyes 
& \cellno 
& \cellno
& \cellno 
& \cellyes  
& \cellno
& \cellno 
& \cellno
& \cellyes
& \cellno
& \cellno 
& \cellno 
& \cellyes
& \cellno
& \cellno 
& \cellno
& \cellno 
& \cellno  
& \cellno  
\\
\hline

& \sysName{$\mathsf{\sf \bf AGNN}$}~\cite{att-gcn2}
&
& \cellyes 
& \cellno 
& \cellno
& \cellno 
& \cellyes  
& \cellno
& \cellno 
& \cellno
& \cellno
& \cellyes
& \cellno 
& \cellno 
& \cellyes
& \cellno
& \cellno 
& \cellno
& \cellno 
& \cellno  
& \cellno  
\\
\hline

& \sysName{$\mathsf{\sf \bf PRML}$}~\cite{path-att-ijcai}
&
& \cellyes 
& \cellno 
& \cellno
& \cellno 
& \cellno 
& \cellyes
& \cellno 
& \cellno
& \cellyes
& \cellno
& \cellno 
& \cellno 
& \cellno
& \cellyes
& \cellno 
& \cellno
& \cellno 
& \cellno 
& \cellno   
\\
\hline

& \sysName{$\mathsf{\sf \bf EAGCN}$}~\cite{node-HIN}
&
& \cellno 
& \cellyes 
& \cellno
& \cellno 
& \cellyes 
& \cellno
& \cellyes 
& \cellno
& \cellyes
& \cellno
& \cellno 
& \cellno 
& \cellno
& \cellno
& \cellno 
& \cellyes
& \cellno 
& \cellno  
& \cellno  
\\
\hline

& \sysName{$\mathsf{\sf \bf Modified-GAT}$}~\cite{mol-att}
&
& \cellyes 
& \cellno 
& \cellno
& \cellno 
& \cellno 
& \cellno
& \cellyes 
& \cellno
& \cellno
& \cellyes
& \cellno 
& \cellno 
& \cellno
& \cellno
& \cellno 
& \cellyes
& \cellno 
& \cellno  
& \cellno  
\\
\hline

& \sysName{$\mathsf{\sf \bf graph2seq}$}~\cite{graph2seq}
&
& \cellyes 
& \cellno 
& \cellno
& \cellno 
& \cellno 
& \cellno
& \cellyes 
& \cellno
& \cellno
& \cellyes
& \cellno
& \cellno  
& \cellno
& \cellno
& \cellno 
& \cellno
& \cellno 
& \cellyes  
& \cellno  
\\
\hline

& \sysName{$\mathsf{\sf \bf GAM}$}~\cite{Lee18}
&
& \cellyes 
& \cellno 
& \cellno
& \cellno 
& \cellno 
& \cellno
& \cellyes 
& \cellno
& \cellyes
& \cellno
& \cellyes 
& \cellno 
& \cellno
& \cellno
& \cellyes 
& \cellno
& \cellno 
& \cellno  
& \cellno  
\\
\hline

& \sysName{$\mathsf{\sf \bf RNNSearch}$}~\cite{Bah15}
&
& \cellno 
& \cellno 
& \cellno
& \cellyes 
& \cellno 
& \cellno
& \cellyes 
& \cellno
& \cellno
& \cellyes
& \cellno 
& \cellno 
& \cellno
& \cellno
& \cellno 
& \cellno
& \cellyes 
& \cellno 
& \cellno  
\\
\hline

& \sysName{$\mathsf{\sf \bf Att-NMT}$}~\cite{Luong15}
&
& \cellno 
& \cellno 
& \cellno
& \cellyes 
& \cellno 
& \cellno
& \cellyes 
& \cellno
& \cellno
& \cellyes
& \cellno
& \cellyes 
& \cellno
& \cellno
& \cellno 
& \cellno
& \cellyes 
& \cellno  
& \cellno  
\\
\hline

& \sysName{$\mathsf{\sf \bf Dipole}$}~\cite{Ma17}
&
& \cellno 
& \cellno 
& \cellno
& \cellyes 
& \cellno 
& \cellno
& \cellyes 
& \cellno
& \cellyes
& \cellyes
& \cellno
& \cellno 
& \cellno
& \cellno
& \cellyes 
& \cellno
& \cellno 
& \cellno  
& \cellno  
\\
\hline

& \sysName{$\mathsf{\sf \bf GRAM}$}~\cite{Choi17}
&
& \cellno 
& \cellno 
& \cellyes
& \cellno 
& \cellno 
& \cellno
& \cellno 
& \cellyes
& \cellyes
& \cellno
& \cellno
& \cellno 
& \cellno
& \cellno
& \cellno 
& \cellno
& \cellno 
& \cellno  
& \cellyes  
\\
\hline

& \sysName{$\mathsf{\sf \bf CCM}$}~\cite{ccm}
&
& \cellyes 
& \cellno 
& \cellno
& \cellno 
& \cellno 
& \cellno
& \cellyes 
& \cellyes
& \cellyes
& \cellno
& \cellno
& \cellno 
& \cellno
& \cellno
& \cellno 
& \cellno
& \cellyes 
& \cellno  
& \cellno  
\\
\hline

& \sysName{$\mathsf{\sf \bf JointD/E + SATT}$}~\cite{Han18}
&
& \cellyes 
& \cellno 
& \cellno
& \cellno 
& \cellno 
& \cellno
& \cellno 
& \cellyes
& \cellno
& \cellyes
& \cellno
& \cellno 
& \cellno
& \cellyes
& \cellno 
& \cellno
& \cellno 
& \cellno  
& \cellno  
\\
\hline

\noalign{\hrule height 0.7pt}
\end{tabular}
\end{center}
}
\end{minipage}
\end{table}
}


%
%
%
%
%
%

\section{Attention-based Node/Edge Embedding} \label{sec:attention-based-node-embedding}
In this section -- as well as the two succeeding sections -- we introduce various graph attention models and categorize them by problem setting. For easy reference, we show all of the surveyed methods in Table~\ref{table:qual-and-quant-comparison}, taking care to highlight where they belong under each of the proposed taxonomies. We now begin by defining the traditional node embedding problem.

\begin{Definition}[Node Embedding] \label{def:node-embedding}
Given a graph $G=(V,E)$ with $V$ as the node set and $E$ as the edge set, the objective of node embedding is to learn a function $f : V \rightarrow \RR^k$ such that each node $i \in V$ is mapped to a $k$-dimensional vector $\vz_i$ where $k \ll |V|$. The node embedding matrix is denoted as $\mZ$.
\end{Definition}

\noindent The learned node embeddings given as output can subsequently be used as input to mining and machine learning algorithms for a variety of tasks such as link prediction~\cite{Sun11}, classification~\cite{Veli18}, community detection~\cite{Newman02}, and role discovery~\cite{roles2015-tkde}. We now define the problem of attention-based node embedding as follows:

\begin{Definition}[Attention-based Node Embedding] \label{def:attention-based-node-embedding}
Given the same inputs as above, we learn a function $f : V \rightarrow \RR^k$ that maps each node $i \in V$ to an embedding vector $\vz_i$. Additionally, we learn a second function $f^{\prime} : V \times V \rightarrow [0, 1]$ to assign ``attention weights'' to the elements in a target node $i$'s neighborhood $\N_i$. The function $f^{\prime}$ defines each neighbor $j$'s, for $j \in \Gamma_i$, importance relative to the target node $i$. Typically, $f^\prime$ is constrained to have $\sum_{j \in \Gamma_i} f^{\prime}(i, j) = 1$ for all $i$ with $f^{\prime}(i, k) = 0$ for all $k \not\in \Gamma_i$. The goal of attention-based node embedding is to assign a similar embedding to node $i$ and to its more similar or important neighbors, \textit{i.e.}, $\delta(\vz_i, \vz_j) > \delta(\vz_i, \vz_k)$ iff $f^{\prime}(i, j) > f^{\prime}(i, k)$, where $\delta$ is some similarity measure.
\end{Definition}

Above, we defined attention-based node embedding; attention-based edge embedding can also be defined similarly. 

Here, we use a single section to discuss both node and edge embeddings since there has not been a lot of work on attention-based edge embeddings and also because the two problems are quite similar~\cite{emb-survey-tkde}. We now categorize the different methods that calculate attention-based node/edge embeddings based on the type of graph they support.

\subsection{Homogeneous graph}
Most of the work that calculate attention-based node embeddings focus on homogeneous graphs~\cite{Veli18, att-node2vec, att-gcn2, path-att-ijcai}. Also, all the methods assume the graph is attributed although~\cite{att-node2vec} only needs node labels (in this case, attribute size $d = 1$).

The method proposed by~\cite{att-node2vec}, called \textsc{AttentionWalks}, is most reminiscent of popular node embedding approaches such as DeepWalk and node2vec~\cite{Perrozi15, node2vec} in that a random walk is used to calculate node contexts. Given a graph $G$ with its corresponding transition matrix $\mT$, and a context window size $c$, we can calculate the expected number of times walks started at each node visits other nodes via:
$$\mathbb{E} [\mD | a_1, \cdots, a_c] = \mI_{n} \sum_{i=1}^{c} a_i (\mT)^i.$$

\noindent Here $\mI_n$ is the size-$n$ identity matrix, $a_i$, for $1 \leq i \leq c$, are learnable attention weights, and $\mD$ is the walk distribution matrix where $D_{uv}$ encodes the number of times node $u$ is expected to visit node $v$ given that a walk is started from each node. In this scenario, attention is thus used to steer the walk towards a broader neighborhood or to restrict it within a narrower neighborhood (\eg, when the weights are top-heavy). This solves the problem of having to do a grid-search to identify the best hyper-parameter $c$ as studies have shown that this has a noticeable impact on performance~\cite{Perrozi15} -- note that for DeepWalk the weights are fixed at $a_i = 1 - \frac{i-1}{c}$. 

Another attention-based method that is very similar in spirit to methods like DeepWalk, node2vec, and LINE~\cite{Perrozi15, node2vec, line} in that it uses co-occurrence information to learn node embeddings is \textsc{GAKE}~\cite{Feng16}. It is important to point out that \textsc{GAKE} builds a graph, commonly referred to as a knowledge graph, from knowledge triplets $(h, t, r)$ where $h$ and $t$ are terms connected by a relationship $r$. Given three triplets (Jose, Tagalog, speaks), (Sato, Nihongo, speaks), and (Jose, Sinigang, eats) we can construct a simple heterogeneous graph with three types of nodes, namely $\{ \text{person, language, food} \}$, and two types of relationships, namely $\{ \text{speaks, eats} \}$. However we categorize \textsc{GAKE} as a homogeneous graph method as it doesn't seem to make a distinction between different kinds of relationships or node types (see metapath2vec~\cite{metapath2vec} for a node embedding method that explicitly models the different kinds of relationships and nodes in a heterogeneous graph). Instead, the method takes a set of knowledge triplets and builds a directed graph by taking each triplet $(h, t, r)$ and adding $h$ and $t$ as the head and tail nodes, respectively, while adding an edge from the head to the tail (they also add a reverse link). The main difference between \textsc{GAKE} and methods such as DeepWalk or node2vec is that they include edges when calculating a node's context. They define three different contexts to get related subjects (nodes or edges), formally they maximize the log-likelihood:
$$\sum_{s \in V \cup E} \sum_{c \in \N_s} log \text{ } \Pr(s\, |\, c)$$

\noindent where $\N_s$ is a set of nodes and/or edges defining the neighborhood context of $s$. To get the final node embedding for a given subject $s$ in the graph, they use attention to obtain the final embedding $\vz_s = \sum_{c \in \N_s^{\prime}}\, \alpha(c) \, \vz_c^{\prime}$ where $\N_s^{\prime}$ is some neighborhood context for $s$, $\alpha(c)$ defines the attention weights for context object $c$ and $\vz_c^{\prime}$ is the learned embedding for $c$. 

On the other hand, methods like \textsc{GAT}~\cite{Veli18} and \textsc{AGNN}~\cite{att-gcn2} extend graph convolutional networks (GCN)~\cite{gcn} by incorporating an explicit attention mechanism. Recall that a GCN is able to propagate information via an input graph's structure using the propagation rule:
$$\mathbf{H}^{(l+1)} = \sigma(\tilde{\mathbf{D}}^{-\frac{1}{2}} \tilde{\mathbf{A}} \tilde{\mathbf{D}}^{-\frac{1}{2}} \mathbf{H}^{(l)} \mathbf{W}^{(l)})$$

\noindent where $\mH^{(l)}$ indicates the learned embedding matrix at layer $l$ of the GCN with $\mH^{(0)} = \mX$. Also, $\tilde{\mathbf{A}} = \mathbf{A} + \mathbf{I}_{n}$ is the adjacency matrix of an undirected graph $\mathbf{A}$ with added self loop. The matrix $\tilde{\mathbf{D}}$, on the other hand, is defined as the diagonal degree matrix of $\tilde{\mathbf{A}}$ so, in other words, $\tilde{D}_{i,i} = \sum_{j} \tilde{A}_{i,j}$. Hence, the term $\tilde{\mathbf{D}}^{-\frac{1}{2}} \tilde{\mathbf{A}} \tilde{\mathbf{D}}^{-\frac{1}{2}}$ computes a symmetric normalization (similar to the normalized graph Laplacian) for the graph defined by $\tilde{\mathbf{A}}$. Finally, $\mathbf{W}^{(l)}$ is the trainable weight-matrix for level $l$ and $\sigma(\cdot)$ is a nonlinearity like ReLU, Sigmoid, or tanh. 

A GCN works like an end-to-end differentiable version of the Weisfeiler-Lehman algorithm~\cite{wl-kernel} where each additional layer allows us to expand and integrate information from a larger neighborhood. However, because we use the term $\tilde{\mathbf{D}}^{-\frac{1}{2}} \tilde{\mathbf{A}} \tilde{\mathbf{D}}^{-\frac{1}{2}} \mathbf{H}$ in the propagation, we are essentially applying a weighted sum of the features of neighboring nodes normalized by their degrees. \textsc{GAT} and \textsc{AGNN} essentially introduce attention weights $a_{uv}$ to determine how much attention we want to give to a neighboring node $v$ from node $u$'s perspective. The two methods differ primarily in the way attention is defined (we expound on this in Sec.~\ref{sec:mechanism}). Furthermore,~\cite{Veli18} introduce the concept of ``multi-attention'' which basically defines multiple attention heads (\eg, weights) between a pair of objects in the graph.

\textsc{PRML}~\cite{path-att-ijcai} is another approach that learns edge embeddings but they use a different strategy to define attention. In~\cite{path-att-ijcai}, a path-length threshold $L$ is defined and a recurrent neural network~\cite{lstm} learns a path-based embedding for paths of length $1, \cdots, L$ between pairs of nodes $u$ and $v$. Attention is defined in two ways. First, attention is used to identify important nodes along paths and this helps in calculating the intermediary path embeddings. Second, attention then assigns priority to paths that are more indicative of link formation and this is used to highlight important or task-relevant path embeddings when these are integrated to form the final edge embedding. We can think of this approach as an attention-based model that calculates the Katz betweenness score for pairs of nodes~\cite{liben-lp}. 

\subsection{Heterogeneous graph}

The only work, to the best of our knowledge, that has been proposed to calculate attention-guided node embeddings for heterogeneous graphs is \textsc{EAGCN}~\cite{node-HIN}. Very similar to both \textsc{GAT} and \textsc{AGNN},~\cite{node-HIN} proposes to apply attention to a GCN~\cite{Duvenaud15, gcn}. However, they handle the case where there can be multiple types of links connecting nodes in a graph. Thus they propose to use ``multi-attention,'' like~\cite{Veli18}, and each of the attention mechanisms considers neighbors defined only by a particular link type. Although the authors validate \textsc{EAGCN} using the graph regression task, the method can be used without much adjustments for node-level tasks since the graph embeddings in \textsc{EAGCN} are simply concatenations or sums of the learned attention-guided node embeddings.

However, the above-mentioned work assumes the given heterogeneous network only has one type of node, \ie, $|\mathcal{T}_{V}| = 1$. 

\subsection{Other special cases}

Unlike the areas of attention-based graph embedding, there does not seem to be work on calculating attention-guided node-embeddings for special types of graphs which appear in certain domains (\eg, medical ontologies represented as DAGs, or certain Heterogeneous networks represented as star graphs).

Since the above-mentioned methods work for general graphs, they should be suitable for more specific types of graphs and one should be able to apply them directly to these cases.

\section{Attention-based Graph Embedding} \label{sec:attention-based-graph-embedding}

Similarly, we begin the discussion by defining the traditional graph embedding problem.

\begin{Definition}[Graph Embedding] \label{def:graph-embedding}
Given a set of graphs, the objective of graph embedding is to learn a function $f : \mathcal{G} \rightarrow \RR^k$ that maps an input graph $G \in \mathcal{G}$ to a low dimensional embedding vector $\vz$ of length $k$; here $\mathcal{G}$ is the input space of graphs. Typically, we want to learn embeddings that group similar graphs (\eg, graphs belonging to the same class) together.
\end{Definition}

\noindent The learned graph embeddings can then be fed as input to machine learning/data mining algorithms to solve a variety of graph-level tasks such as graph classification~\cite{Lee18}, graph regression~\cite{Duvenaud15}, and graph-to-sequence generation~\cite{graph2seq}. We now define the problem of attention-based graph embedding as follows:

\begin{Definition}[Attention-based Graph Embedding] \label{def:attention-based-graph-embedding}
Given the same inputs as above, we learn a function $f : \mathcal{G} \rightarrow \RR^k$ that maps each input graph $G \in \mathcal{G}$ to an embedding vector $\vz$ of length $k$. Additionally, we learn a second function $f^{\prime}$ that assigns ``attention weights'' to different subgraphs of a given graph to allow the model to prioritize more important parts (\ie, subgraphs) of the graph when calculating its embedding.
\end{Definition}

\subsection{Homogeneous graph}
Several methods have been proposed for learning graph embeddings on homogeneous graphs~\cite{Lee18, graph2seq, mol-att}.

\cite{mol-att} simply use the formulation of \textsc{GAT}~\cite{Veli18} and make a few adjustments to the way attention is calculated. However, aside from the change in how attention weights are calculated, there isn't much difference between \textsc{GAT} and the method proposed in~\cite{mol-att}. To obtain the final graph embedding for the task of graph regression on molecular graphs, the proposed method adds fully connected layers after the final GAT layer to flatten the per-node outputs.

On the other hand,~\cite{graph2seq} propose \textsc{graph2seq} which solves the natural language question answering task. Given a set of facts, in the form of sentences their approach is quite unique in that they convert the input into an attributed directed homogeneous graph. To obtain the graph embedding that is used for the sequence generation task, they first learn node embeddings for each node in the graph. Node embeddings are formed by concatenating a forward and a backward representation for each node which are representations derived by aggregating information from each node's forward (neighbors traversed using forward links) and backward (similarly, neighbors traversed using reverse links) neighborhoods, respectively. The final graph embedding is obtained by pooling the individual node embeddings or by simply aggregating them. Attention, in the case of \textsc{graph2seq} however, is applied by also attending to the individual node embeddings during sequence generation. Since each node embedding $\vz_i$ captures information in the region around node $i$ (we can think of this as a subgraph focused around $i$), attention allows use to prioritize a particular part of the graph when generating a word in the output sentence (sequence). This captures the intuition that different parts of the graph can be associated, primarily, with different concepts or words.

Finally, in a previous work~\cite{Lee18}, we proposed \textsc{GAM} which uses two types of attention to learn a graph embedding. The main idea is to use an attention-guided walk to sample relevant parts of the graph to form a subgraph embedding. In \textsc{GAM}, we took a walk of length $L$. Let $1 \leq i \leq L$ be the $i$-th node discovered in the walk and $\vx_i$ the corresponding node attribute vector, an RNN is used to integrate the node attributes ($\vx_1, \cdots, \vx_L$) to form a subgraph embedding $\vs$ (the subgraph or region covered during the walk). During each step, an attention mechanism is used to determine which neighboring node is more relevant to allow the walk to cover more task-relevant parts of the graph. To get the final graph embedding, we deploy $z$ ``agents'' to sample from various parts of the graph yielding embeddings $\{\vs_1, \cdots, \vs_z\}$. A second attention mechanism is then used to determine the relevance of the various subgraph embeddings before these are combined to form a graph embedding. In other words an attention mechanism is defined as a function $\alpha : \mathbb{R}^{k} \rightarrow [0,1]$ which maps a given subgraph embedding to a relevance score. The graph embedding is thus defined as $\sum_{i=1}^{z} \alpha(\vs_i) \, \vs_i$.

\subsection{Heterogeneous graph}
Recall from our previous discussion that \textsc{EAGCN}~\cite{node-HIN} was used to study the task of graph regression. \textsc{EAGCN} used an approach similar to \textsc{GAT} to generate a graph embedding. Since the method uses a similar attention mechanism as \textsc{GAT}, attention is focused on determining important neighbors to attend to when calculating node embeddings for a graph. The final graph embedding is then a concatenation/sum of the attention-guided node embeddings. It shouldn't be difficult to apply a second attention mechanism, similar to that of~\cite{Lee18}, to weight the importance of the different node embeddings. 

\subsection{Other special cases}
Attention was originally studied in the Computer Vision and NLP domains and various RNN models using attention on sequence-based tasks were proposed~\cite{Bah15, Luong15}. Since sequences are technically no different from paths, we introduce some notable attention models under this setting.

A sequence (\eg, a sentence) of length $L$ can be represented as a directed attributed graph of length $L$ -- the $i$-th attribute vector $\vx_i$ of node $i$ is then a representation of the $i$-th component in the sequence (a one-hot word embedding or a word2vec embedding, for instance). In the proposed methods~\cite{Bah15, Luong15, Ma17}, the node attribute vectors $\{ \vx_i, \cdots, \vx_L \}$ are fed one after the other into an RNN. Recall that in the simplest case, a recurrent neural network calculates a hidden embedding $\vh_i$ for each input $i$ via the rule 
$$\vh_i = \sigma(\mW_h\vx_i + \mU_h\vh_{i-1}) + \vb_h,$$

\noindent where $\mW_h$ and $\mU_h$ are trainable weight matrices for the input $\vx_i$ and the previous hidden embedding $\vh_{i-1}$, respectively; $\vb_h$ is a bias vector and $\vh_0$ is usually initialized to the zero vector.

In this case, we can think of $\vh_i$ as node $i$'s node embedding which integrated information from the sub-path encompassing nodes $1, \cdots, i$. Attention can then be applied on node embeddings to generate an attention-guided graph embedding. In particular, the method proposed in~\cite{Bah15} called \textsc{RNNSearch} defined the graph embedding as $\vz = \sum_{i=1}^{L} \alpha_i \, \vh_i$ where attention is assigned to each hidden embedding $\vh_i$ depending on its relevance to the task. 

When the length of the path $L$ is large, however, attending over all the hidden embeddings as in \textsc{RNNSearch} may not yield the best results.~\cite{Luong15} proposed to use two attention mechanisms, the first is similar in spirit to that of~\cite{Bah15}. However, the second attention allows the model to select a local point within the input and focus attention there. More concretely, depending on the needs of the task, the second attention mechanism outputs a position $1 \leq p \leq L$ and the graph embedding is constructed from the hidden node embeddings $\vh_{p-D}, \cdots, \vh_{p+D}$ where $D$ is an empirically selected attention window size.

Finally, \textsc{Dipole}~\cite{Ma17} applied an attention model similar to that of~\cite{Bah15} to sequential medical data. They used it to diagnose or predict medical conditions from the sequence. In practice, both~\cite{Ma17} and~\cite{Bah15} used a bi-directional model which processes an input sequence using two RNNs, one taking the input sequence in its original order and another which takes the input sequence in reverse order.

\section{Attention-based Hybrid Embedding} \label{sec:attention-based-hybrid-embedding}
In this section we discuss methods that apply attention on graph data. However, the calculated embeddings are ``hybrid'' since the methods here also take data of other modalities (\eg, text) and the learned embedding is a combination of all inputs.

\subsection{Homogeneous graph}
Like \textsc{GAKE}~\cite{Feng16}, \textsc{CCM}~\cite{ccm} deals with knowledge graphs. However, in~\cite{ccm}, the sequence-to-sequence generation problem is studied. \textsc{CCM} uses an encoder-decoder model~\cite{encoder-decoder} to encode a sentence and output a sentence (\eg, for sentence translation, or question and answering). However, the setting used in \textsc{CCM} assumes that the model refers to a knowledge graph for the question-and-answer task. Their knowledge graph is made up of multiple connected components, each of which is made up of multiple knowledge triplets. A subgraph embedding $\vz_i$ is learned for each component $i$:

$$\vz_i = \sum_{n=1}^{|T_i|} \alpha_n^{(i)} [\vh_n^{(i)};\vt_n^{(i)}]$$

\noindent where $T_i = \{ (\vh_1^{(i)}, \vt_1^{(i)}, \vr_1^{(i)}), \cdots,  (\vh_{n_i}^{(i)}, \vt_{n_i}^{(i)}, \vr_{n_i}^{(i)}) \}$ consists of the embeddings for the corresponding knowledge triplets for $i$ and $\alpha_n^{(i)}$ is used to assign importance to different triplets (represented by the concatenation of the term embeddings) taking into account the terms as well as their relationship.

While the model is in the process of decoding (generating words for the output), it selectively identifies important subgraphs by attending on their embeddings $\vz_i$. Furthermore, an attention mechanism is also used to identify the triplets within a selected knowledge subgraph to identify important triplets for word generation. Like \textsc{GAKE}, we consider \textsc{CCM} a homogeneous graph method since it doesn't seem to make an explicit distinction between different types of nodes and relationships.

\textsc{JointD/E + SATT}~\cite{Han18} is another work that applies attention on a knowledge graph, similar to~\cite{Feng16, ccm}. However, they also use a large text corpus that may contain references to the different terms in the knowledge graph implying certain relationships. They introduce a method that uses attention to learn a joint embedding from graph and text data for the task of knowledge graph link prediction.

Under certain settings (brain network construction, for instance) the datasets tend to be quite small and noisy~\cite{sdm16zhang}. Methods such as that proposed by~\cite{sdm16zhang} proposes to use side information to regularize the graph construction process to highlight more discriminative patterns -- which is useful when the output graphs are used for graph classification. Exploring the possibility of adding attention in this setting is interesting.

\subsection{Heterogeneous graph}
While there has been work proposed like~\cite{Han18, ccm} that applies attention on knowledge graphs -- which are considered heterogeneous graphs -- these models do not distinguish between the different types of links and nodes explicitly. To the best of our knowledge, there currently does not exist any work that has considered this setting. One possibility is to extend the idea proposed by \textsc{EAGCN}~\cite{node-HIN} to more general heterogeneous graphs.

\subsection{Other special cases}
\textsc{GRAM}~\cite{Choi17} is a graph-based attention model for doing classification/regression on clinical data. Clinical records can usually be described by clinical codes $c_1, \cdots, c_{|\mathcal{C}|} \in \mathcal{C}$ in a vocabulary $\mathcal{C}$. \textsc{GRAM} constructs a DAG whose leaves are the codes $c_1, \cdots, c_{|\mathcal{C}|}$ while the ancestor nodes are more general medical concepts.~\cite{Choi17} uses an attention mechanism to learn a $k$-dimensional final embedding of each leaf node $i$ (medical concept) via:

$$\vg_i = \sum_{j \in \mathcal{A}(i)} \alpha_{i,j} \, \ve_j.$$

\noindent where $\mathcal{A}(i)$ denotes the set comprised of $i$ and all its ancestors and $\ve_j$ is the $k$-dimensional basic embedding of the node $j$. The use of attention allows the model to refer to more informative or relevant general concepts when a concept in $\mathcal{C}$ is less helpful for medical diagnosis (\eg, it is a rare concept). Since a patient's clinical visit record is represented as a binary vector $\vx \in \{0,1\}^{|\mathcal{C}|}$ indicating which clinical codes were present for a particular visit the learned embeddings $\vg_1, \cdots, \vg_{|\mathcal{C}|}$ can be stacked to form a $k \times |\mathcal{C}|$ embedding matrix to embed $\vx$; this can then be inputted into a predictive model for medical diagnosis. Note that the problem of medical diagnosis is a classical supervised learning task which takes clinical code feature vectors but \textsc{GRAM} applies attention on a medical concept DAG for the purpose of embedding the given feature vectors.

\section{Types of Graph Attention Mechanism} \label{sec:mechanism}
We now describe the three main types of attention that have been applied to graph data. While all three types of attention share the same purpose or intent, they differ in how attention is defined or implemented. In this section we provide examples from the literature to illustrate how each type is implemented.

Recall from our general definition of attention in Def.~\ref{def:graph-attention} that we are given a target graph object (\eg, node, edge, graph, \textit{etc}) $v_0$ and a set of graph objects in $v_0$'s ``neighborhood'' $\{v_1, \cdots, v_{|\N_{v_0}|} \} \in \N_{v_0}$. Attention is defined as a function $f^\prime : \{ v_0 \} \times \N_{v_0} \rightarrow [0, 1]$ that maps each of the objects in $\N_{v_0}$ to a relevance score. In practice this is usually done in one of three ways which we introduce below. 

\begin{figure}[t]
\centering
\includegraphics[width=0.9\linewidth]{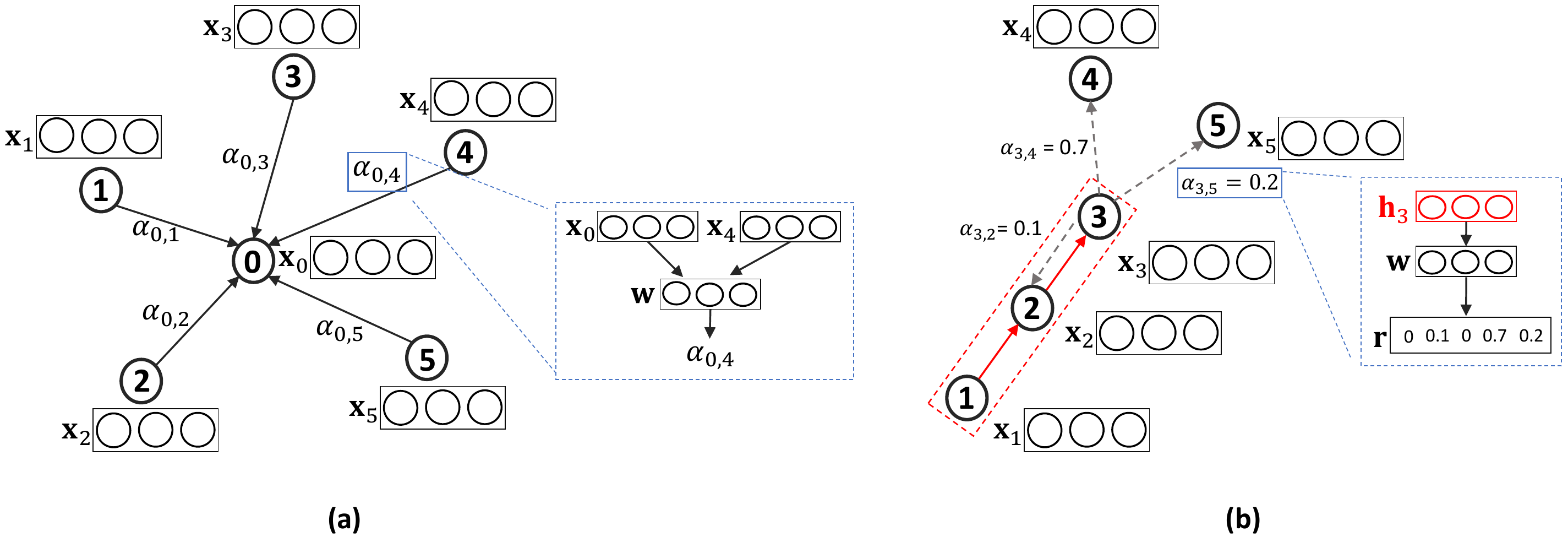}
\caption{(a) Given a target object $v_0$, we assign importance weights $\alpha_{0, i}$ to the objects $i$ in our neighborhood. This can be done by learning a function that assigns importance by examining (possibly) the hidden embeddings of $v_0$ and $v_i$, $\vx_0$ and $\vx_i$. (b) The hidden embedding $\vh_3$ represents information ($\vx_1, \cdots, \vx_3$) we've integrated after a walk of length $L=3$, we input this into a ranking function that determines the importance of various neighbor nodes and this is used to bias the next step. For both examples, we use $\vec{w}$ to represent the trainable parameters of the attention function.}
\label{fig:att-diff}
\vspace{-5mm}
\end{figure}

\subsection{Learn attention weights}
Given the corresponding attributes/features $\vx_{0}, \vx_{1}, \cdots, \vx_{|\N_{o^*}|}$ for $v_0, v_1, \cdots, v_{|\N_{v_0}|}$, attention weights can be learned via:

$$\alpha_{0,j} = \frac{e_{0,j}}{\sum_{k \in \N_{v_0}} e_{0, k}}$$

\noindent where $e_{0,j}$ is node $v_j$'s relevance to $v_0$. In practice, this is typically implemented using softmax with a trainable function learning $v_j$'s relevance to $v_0$ by considering their attributes. The implementation in \textsc{GAT}~\cite{Veli18} illustrates this:

$$\alpha_{0,j} = \frac{\text{exp} \Big( \text{LeakyReLU} \Big( \va [\mW \vx_0||\mW \vx_j] \Big) \Big)}{\sum_{k \in \N_{v_0}} \text{exp} \Big( \text{LeakyReLU} \Big( \va [\mW \vx_0||\mW \vx_k] \Big) \Big)}$$

\noindent where $\va$ is a trainable attention vector, $\mW$ is a trainable weight matrix mapping the input features to the hidden space, and $||$ represents concatenation. An illustration of this is shown in Fig.~\ref{fig:att-diff}a.

\subsection{Similarity-based attention}
Again, given the corresponding attributes or features, the second type of attention can be learned similarly as above except for a key difference. We call this approach similarity-based attention as more attention is given to object's that share more similar hidden representations or features, this is also often referred to in the literature as alignment~\cite{Bah15}. To illustrate this, we use the definition given in \textsc{AGNN}~\cite{att-gcn2}:

$$\alpha_{0,j} = \frac{\text{exp} \Big(\beta \cdot \text{cos} \Big( \mW \vx_0, \mW \vx_j \Big) \Big)}{\sum_{k \in \N_{v_0}} \text{exp} \Big(\beta \cdot \text{cos} \Big( \mW \vx_0, \mW \vx_k \Big) \Big)}$$

\noindent where $\beta$ is a trainable bias and ``cos'' represents cosine-similarity; like before, $\mW$ is a trainable weight matrix to map inputs to the hidden space. Note that this is very similar to the above definition. The difference is that the model explicitly learns similar hidden embeddings for objects that are relevant to each other since attention is based on similarity or alignment.

\subsection{Attention-guided walk}
While the first two types of attention focuses on choosing relevant information to integrate into a target object's hidden representation, the third type of attention has a slightly different purpose. We use \textsc{GAM}~\cite{Lee18} to illustrate this idea. \textsc{GAM} takes a series of steps on an input graph and encodes information from visited nodes using an RNN to construct a subgraph embedding. The RNN's hidden state at a time $t$, $\vh_t \in \mathbb{R}^{h}$ encodes information from the nodes that were previously visited by the walk from steps $1, \cdots, t$. Attention is then defined as a function $f^\prime : \mathbb{R}^{h} \rightarrow \mathbb{R}^{k}$ that maps an input hidden vector $f^\prime(\vh_t) = \vr_{t+1}$ to a $k$-dimensional rank vector that tells us which of the $k$ types of nodes we should prioritize for our next step. The model will then prioritize neighboring nodes of a particular type for the next step. We illustrate this in Fig.~\ref{fig:att-diff}b.

%
%
%
%

\section{Graph Attention Tasks} \label{sec:tasks}
The different attention-based graph methods can be divided broadly by the kind of problem they solve: node-level or graph-level.

\subsection{Node-level}
A number of work have been proposed to study graph attention for node-level tasks, the most notable of which are node classification and link prediction~\cite{Veli18, att-node2vec, att-gcn2, path-att-ijcai, Feng16, Han18}. Although each method differs in their approach and assumptions, they share a common technique which is to learn an attention-guided node or edge embedding which can then be used to train a classifier for classification or link prediction. It isn't hard to see these methods implemented for the related tasks of node clustering. Some notable node-level tasks for which attention-based graph methods have not been proposed include node/edge role discovery~\cite{roles2015-tkde}, and node ranking~\cite{rankclus}.

\subsection{Graph-level}
Multiple works have also studied graph-level tasks such as graph classification and graph regression. In this setting, an attention-guided graph embedding is constructed by attending to relevant parts of the graph. Methods like \textsc{EAGCN}, \textsc{Modified-GAT}, \textsc{GAM}, and \textsc{Dipole}~\cite{Ma17, Lee18, node-HIN, mol-att} learn a graph embedding for the more standard graph-based tasks of classification and regression. It is not hard to see how these methods can be applied to the related problem of graph similarity search~\cite{simsearch}. 

On the other had, work like~\cite{graph2seq, Bah15, Luong15, ccm} generate sequences from input graph data. Notably, \textsc{graph2seq} proposes a method that outputs a sequence given an input graph instead of the more methods that do sequence-to-sequence generation.

Finally, \textsc{GRAM}~\cite{Choi17} applied attention to a medical ontology graph to help learn attention-based embeddings for medical codes. While the problem they studied was the problem of classifying a patient record (described by certain medical codes), the novelty of their work was in applying attention on the ontology graph to improve model performance.

%
%

\section{Discussion and Challenges} \label{sec:discussion-challenges} 
In this section we discuss additional issues and highlight important challenges for future work.

\subsection{Attention-based methods for heterogeneous graphs}
The study of heterogeneous graphs, also called heterogeneous information networks, has become quite popular in recent years with many papers published in the area~\cite{Sun11, metapath2vec, rankclus, linklee, rankclusstar}. 

While methods like \textsc{GAKE}, \textsc{CCM}, \textsc{JointD/E+SATT} all consider knowledge graphs which is a type of heterogeneous graph, they do not differentiate between the different kinds of links and nodes. While methods such as \textsc{EAGCN} deal with multiple types of links they do not consider the general case where there can also be multiple kinds of nodes. \textsc{GAM} is another method that distinguishes, in a way, between different kinds of nodes since the attention-guided walk prioritizes the node to visit based on node type.

There is a need for attention-based methods that study how attention can be applied to general heterogeneous networks, especially taking into consideration how different kinds of meta-paths~\cite{metapath2vec} can affect the learned embeddings. This is important as methods based on heterogeneous graphs have been shown to outperform methods that make the assumption that graphs only have a single type of link/edge. One can refer to~\cite{Shi17} for a survey of heterogeneous graph-based methods.

\subsection{Scalability of graph attention models}
The majority of methods~\cite{Bah15, ccm, Ma17} that calculate an attention-based graph embedding work for relatively small graphs and may have trouble scaling effectively to larger graphs. 

Methods like~\cite{mol-att, node-HIN, Lee18} may be able to work on larger graphs but they still have their shortcomings. For instance, \textsc{GAM}~\cite{Lee18} uses a walk to sample a graph. For large graphs, natural questions that arise include: (1) Do we need longer walks and can an RNN handle these? (2) Can a walk effectively capture all relevant and useful structures, especially if they are more complex?

Methods like~\cite{mol-att, node-HIN} that apply attention to a GCN architecture seem like a step in the right direction as this can be described as an attention-based end-to-end differentiable version of the Weisfeiler-Lehman algorithm~\cite{Duvenaud15, wl-kernel} especially if we train stochastically~\cite{stoc-gcn}. However, there is a need to evaluate graph attention models on a variety of large real-world graphs to test their effectiveness. It would also be useful to explore other ways to apply attention to improve performance.


\subsection{Inductive graph learning}
Recently, there has been an interest in exploring inductive learning of graph-based problems~\cite{Hamilton17, Guo18} which is different from transductive learning. The former is more useful than the latter as it can generalize to yet unseen data. 

This setting is important as it allows graph-based methods to handle many real-world cases where nodes and edges in graphs can appear dynamically. It also allows us to learn general patterns in one graph dataset that can be useful in another dataset much like how transfer learning is used to train a model on one text corpora for application on another text dataset~\cite{Dai07} or to train a model to recognize general shapes in a large image dataset to be applied to a sparser dataset~\cite{Oquab14}. Another interesting example of transfer learning is shown by~\cite{Zhu14} where the transfer is done across data of different domains (\eg, text to images). 

An interesting direction for future study is looking at attention-based inductive learning techniques that can be used to identify relevant graph patterns that are generalizable to other graph datasets. Looking further, we can also explore attention-based techniques that do cross-domain or heterogeneous transfer learning~\cite{Zhu14}.

While the authors of methods like \textsc{GAT}~\cite{Veli18} have conducted an initial study of inductive learning on a small graph dataset, we believe more focused experiments should be done on graph-based inductive learning taking into account a large set of datasets and settings.

\subsection{Attention-guided attributed walk} \label{attention-guided-attr-walk}
Recently, Ahmed~\etal~\cite{role2vec} proposed the notion of \emph{attributed walk} and showed that it can be used to generalize graph-based deep learning and embedding methods making them more powerful and able to learn more appropriate embeddings.
This is achieved by replacing the notion of random walk (based on node ids) used in graph-based deep learning and embedding methods with the more appropriate and powerful notion of \emph{attributed walk}.
More formally, 
\begin{Definition}[Attributed walk]\label{def:attr-walk}
Let $\vx_i$ be a $k$-dimensional vector for node $v_i$.
An \emph{attributed walk} $S$ of length $L$ is defined as a sequence of adjacent node types 
\begin{gather} \label{eq:attr-random-walk}
\phi(\vx_{i_{1}}), \phi(\vx_{i_{2}}),\ldots, \phi(\vx_{i_{L+1}})
\end{gather}
associated with a sequence of indices $i_{1}, i_{2}, \ldots, i_{L+1}$ such that $(v_{i_{t}}, v_{i_{t+1}}) \in E$ for all $1 \leq t \leq L$ and
$\phi : \vx \rightarrow y$ is a function that maps the input vector $\vx$ of a node to a corresponding type $\phi(\vx)$.
\end{Definition}
The type sequence $\phi(\vx_{i_{1}}), \phi(\vx_{i_{2}}),\ldots, \phi(\vx_{i_{L+1}})$ is the node types that occur during a walk (as opposed to the node ids).
%
It was shown that the original deep graph learning models that use traditional random walks are recovered as a special case of the attributed walk framework when the number of unique types $t \rightarrow n$~\cite{role2vec}. It should be noted that the node types here do not necessarily refer to node types in heterogeneous graphs but can be calculated for nodes in a homogeneous graph from their local structure.
%

Attention-based methods that leverage random walks (based on node ids) may also benefit from the notion of \emph{attributed walks} (typed walks) proposed by Ahmed~\etal\cite{role2vec}.

\section{Conclusion} \label{sec:conclusion}
In this work, we conducted a comprehensive and focused survey of the literature on the important field of graph attention models. To the best of our knowledge, this is the first work of this kind. We introduced three intuitive taxonomies to group existing work. These are based on problem setting, the type of attention mechanism used, and the task. We motivated our taxonomies through detailed examples and used each to survey competing approaches from the taxonomy's unique standpoint. We also highlighted several challenges in the area and provided discussion on possible directions for future work.

\bibliographystyle{ACM-Reference-Format}
\bibliography{paper}

\end{document}